\def\year{2022}\relax
\documentclass[letterpaper]{article} 
\usepackage{aaai22}  
\usepackage{times}  
\usepackage{helvet}  
\usepackage{courier}  
\usepackage[hyphens]{url}  
\usepackage{graphicx} 
\usepackage{natbib}  
\usepackage{caption} 
\DeclareCaptionStyle{ruled}{labelfont=normalfont,labelsep=colon,strut=off} 
\usepackage{bm}
\usepackage{amsmath}
\usepackage{amsfonts}
\usepackage{amssymb}
\usepackage{algorithm}
\usepackage{subfigure}
\usepackage{diagbox}
\usepackage{algpseudocode}
\usepackage{multirow}
\usepackage[smallscripts]{moresize}

\def\1{\mathbf{1}}
\def\0{\mathbf{0}}

\def\x{{\bf x}}

\def\K{{\bf K}}

\def \aalpha {\bm{\alpha}}

\def\z{{\bf z}}

\newtheorem{assumption}{\bf Assumption}
\newtheorem{remark}{\bf Remark}
\newtheorem{definition}{\bf Definition}
\newtheorem{lemma}{\bf Lemma}
\newtheorem{theorem}{\bf Theorem}

\newtheorem{proposition}{\bf Proposition}

\newcommand{\smalldisplayskips}{%
	\setlength{\abovedisplayskip}{4pt}%
	\setlength{\belowdisplayskip}{4pt}%
	\setlength{\abovedisplayshortskip}{4pt}%
	\setlength{\belowdisplayshortskip}{4pt}}

\newenvironment{proof}{\hspace{0ex}\textsc{Proof}.\hspace{1ex}}{\hfill$\blacksquare$\newline}
\newcommand*\dif{\mathop{}\!\mathrm{d}}

\title{Regularized Modal Regression on Markov-Dependent Observations: A Theoretical Assessment}
\author{
	Tieliang Gong\textsuperscript{\rm 1,2},
	Yuxin Dong\textsuperscript{\rm 1,2},
	Hong Chen\textsuperscript{\rm 3},
	Wei Feng\textsuperscript{\rm 1,2},
	Bo Dong\textsuperscript{\rm 2,4},
	Chen Li\textsuperscript{\rm 1,2}
}
\affiliations{
	\textsuperscript{\rm 1}School of Computer Science and Technology,
	Xi'an Jiaotong University, Xi'an 710049, China\\
	\textsuperscript{\rm 2}Key Laboratory of Intelligent Networks and Network Security, Ministry of Education, Xi'an 710049, China\\
	\textsuperscript{\rm 3}College of Science, Huazhong Agriculture University, Wuhan 430070, China\\
	\textsuperscript{\rm 4}School of Continuing Education, Xi'an Jiaotong University, Xi'an,710049, China\\
	adidasgtl@gmail.com, dongyuxin@stu.xjtu.edu.cn, chenh@mail.hzau.edu.cn, weifeng.ft@foxmail.com, dong.bo@mail.xjtu.edu.cn, cli@xjtu.edu.cn
}

\begin{document}

\maketitle

\begin{abstract}
	Modal regression, a widely used regression protocol,  has been extensively investigated in statistical and machine learning communities due to its robustness to outliers and heavy-tailed noises. Understanding modal regression's theoretical behavior  can be fundamental in learning theory. Despite significant progress in characterizing its statistical property, the majority of the results are based on the assumption that samples are independent and identical distributed (i.i.d.), which is too restrictive for real-world applications. This paper concerns the statistical property of regularized modal regression (RMR) within an important dependence structure - Markov dependent. Specifically, we establish the upper bound for RMR estimator under moderate conditions and give an explicit learning rate. Our results show that the Markov dependence impacts on the generalization error in the way that sample size would be discounted by a multiplicative factor depending on the spectral gap of underlying Markov chain. This result shed a new light on characterizing the theoretical underpinning for robust regression. 
\end{abstract}

\section{Introduction}\label{sec:intro}
In this paper, we consider the non-parametric regression problem which aims at inferring the relationship between input and output.  To formulate this problem, denote  $X$ as the covariate variable that takes values in a compared metric space $\mathcal{X} \subset \mathbb{R}^d$ and $Y$ that take values in $\mathcal{Y} = \mathbb{R}$. The sample pair $(X, Y)$ is generated from the following model :
\begin{equation*}
Y = f^*(X) + \epsilon,
\end{equation*}
where $\epsilon$ is the noise term.  The goal of non-parametric regression is to find the unknown function $f^\star$ in a non-parametric manner while some certain assumptions on noise term are imposed. This problem can be boiled down to learn a characterization of the conditional distribution, given  a set of observations. Some commonly used characterizations include the conditional mean \cite{tibshirani1996regression}, the conditional quantile \cite{yu2003quantile,meinshausen2006quantile} and the conditional mode \cite{chen2016nonparametric,feng2020learning}, which correspond to mean regression, quantile regression and modal regression (MR), respectively. Each regression protocol has its own benefits in modeling the noise. For instance, conditional mean regression can achieve satisfactory effect when the noise is Gaussian or sub-Gaussian, while regression towards the conditional quantile and conditional mode can be more robust in complex noise cases. In practice, selecting an appropriate regression protocol usually depends on the data type.

Modal regression is an appropriate regression protocol when facing heavy-tailed noises and outliers. Different from the conventional mean regression, which aims to estimate the conditional mean, modal regression seeks for the unknown truth $f^\star$ by regressing towards to the conditional mode function. For a set of observations, the mode denotes the value that appears most frequently. In the context of density estimation, the mode is the value at which the density function achieves its peak value. Hence, conditional mode can reveal the structure of outputs and the trends of observations. Research on modal regression can be broadly classified into two categories:  (semi-) parametric and nonparametric approaches. For parametric approaches, a parametric form of the global conditional mode function is required. To name a few, studies in \cite{lee1989mode,yu2012bayesian,yao2014new,lv2014robust,khardani2017non} fall in this setting. For non-parametric approaches, the conditional mode is sought by maximizing a conditional density or a joint density which is typically estimated in a non-parametric manner.  Typical works include \cite{chen2016nonparametric,feng2020learning,yao2016nonparametric,zhou2016nonparametric,wang2017regularized}.  Great progress on understanding  the theoretical property of modal regression estimator  have been made during the last two decades (we  refer the reader to \cite{feng2020learning}). In particular, Chen et al.  \cite{chen2016nonparametric} derived asymptotic error bounds for local modal regression  within the framework of kernel density estimation.  Feng et al.  \cite{feng2020learning} established the statistical consistency for  modal regression estimator by assuming the existence of  global conditional  mode function.

All the works mentioned above are based on the assumption that data are independent and identical distributed (i.i.d.). Nevertheless, this assumption is too restrictive in a broad range of  real datasets. As a matter of fact, a considerable number of real datasets are tempera in nature. 
For example, functional magnetic resonance imaging (fMRI) data \cite{ryali2012estimation,smith2012future} are usually collected from different regions over a time period; the macroeconomic data \cite{mccracken2016fred} span the time periods of decades and are kept updating till now.  It poses challenges for researchers applying modal regression to these time-series data. Therefore, understanding the statistical behavior of modal regression estimator for time-series data can be one of the most important issues. 

This paper aims to close the gap between theories of modal regression and practical requirements in addressing dependent observation of real data. Inspired by the statistical guarantees of modal regression dealing with heavy-tailed errors in the independent setup, we consider  extending MR to cope with dependent structure of observations. Albeit  convergence rates on modal regression are given in \cite{feng2020learning,wang2017regularized}, it is still unclear whether these results work for dependent observations. As an initial exploration on this topic, this paper narrows down to Markov chain, an important and  widely used dependence structure, investigating the generalization performance of regularized modal regression (RMR) on Markov-dependence data. Within the Markov-dependence setup, we first show that RMR estimator is statistical consistent under moderate conditions, and establish its explicit convergence rates with order $\mathcal{O} \big( (1 -\gamma^2)^{-\frac{1}{5}} m^{-\frac{1}{5}}  \big)$ under appropriate parameter selection, where $m$ is number of Markov-dependent observations,  and $\gamma$ is the absolute spectral gap of the underlying Markov chain.  

The rest of the paper is organized as follows. Section 2 introduces the necessary notions and notations. Section 3  presents the assumptions and the main theorems. Section 4 sketches the proofs of the main theorems. Finally, a brief discussion is concluded in Section 5.

\section{Model and Methodology} 
\label{sec:pre}
\subsection{Model Setup}
 Let $\mathcal{X} \in \mathbb{R}^d$ and $\mathcal{Y} \in \mathbb{R}$ be the input and output spaces respectively. In the modal regression setting,  training samples $\z = \{ (\x_i, y_i)\}_{i=1}^{m} \subset \mathcal{X} \times \mathcal{Y}$ are generated independently by
\begin{equation}\label{eq_model}
Y = f^{*}(X) + \epsilon,
\end{equation}
where the mode of the conditional distribution of $\epsilon$ at any $\x \in \mathcal{X}$ is assumed to be zero, i.e.
\begin{equation}
	\small
\textbf{mode}(\varepsilon| X=\x) = \arg \max_{t} p_{\epsilon |X}(t| X= \x) = 0,~ \forall \x \in \mathcal{X},
\end{equation}
where $p_{\epsilon|X}$ be the conditional density of $\varepsilon$ on $X$. Then, the target function of modal regression can be represented by
\begin{equation} \label{eq_mode}
	\small
f^*(\x) =  \textbf{mode}(Y| X=\x) = \arg \max_{t} p_{Y|X} (t| X = \x).
\end{equation} 
Throughout this paper, we assume that  for any $\x \in \mathcal{X}$, $ \arg \max_{t} p_{Y|X} (t| X = \x)$ is well defined, which is equivalent to the existence and uniqueness of the global mode of the conditional density $p_{Y|X}$.  Moreover, we assume that $f^*$ is bounded, i.e $\|f^*\|_\infty \leq M$ for some $M>0$. 

Denote $\rho$ on $\mathcal{X} \times \mathcal{Y}$ as the intrinsic distribution for data generated by (\ref{eq_model}) and denote $\rho_{\mathcal{X}}$ as the corresponding marginal distribution on $\mathcal{X}$.  For any measurable function $f: \mathcal{X} \rightarrow \mathbb{R}$, the modal regression performance can be characterized by
\begin{equation} \label{eq_modal_risk}
\smalldisplayskips
\mathcal{R}(f) = \int_{\mathcal{X}}p_{Y|X} \big(f(\x)| X = \x \big) \dif \rho_{\mathcal{X}}(\x).
\end{equation}
It has been proved that $f^*$ is the maximizer of (\ref{eq_modal_risk}) over all measurable functions \cite{feng2020learning}. Since $\rho_{\mathcal{X}}$ and $p_{Y|X}$ are usually unknown, we can not calculate the estimator directly by maximizing (\ref{eq_modal_risk}). Feng et al.  \cite{feng2020learning} proved $\mathcal{R}(f) = p_{\epsilon_f}(0)$, where $p_{\varepsilon_f}$ is the density function of random variable $\epsilon_f = Y - f(X)$. This implies that maximizing $\mathcal{R}(f)$ over some hypothesis spaces is equivalent to maximizing the density of $\varepsilon_f$ at $0$, which can be estimated by non-parametric kernel density estimation.

Let $K_\sigma: \mathbb{R} \times \mathbb{R} \rightarrow \mathbb{R}_+$ be a kernel function, and $\phi(\frac{u - u^\prime}{\sigma}) = K_\sigma(u, u^\prime)$ be a representing function which satisfies $\phi(u) = \phi(-u)$, $\phi(u) \leq \phi(0)$ for any $u \in \mathbb{R}$ and $\int_{\mathbb{R}} \phi(u) \dif u= 1$. With the help of $K_\sigma$, we can obtain the empirical estimation of $\mathcal{R}(f)$ by kernel density estimation, given by
\begin{equation*}
\tiny
\smalldisplayskips
\mathcal{R}_{\z}^{\sigma} \!(f) \!=\! \frac{1}{m \sigma} \!\sum_{i=1}^{m}\! K_\sigma(y_i \!-\! f(\x_i), 0) \!=\! \frac{1}{m \sigma} \! \sum_{i=1}^{m} \!\phi \Bigg(\! \frac{y_i \!-\! f(\x_i)}{\sigma} \! \Bigg).
\end{equation*}
For any $f: \mathcal{X} \rightarrow \mathbb{R}$, the expectation version of $\mathcal{R}_{\z}^{\sigma} (f)$ is
\begin{equation*}
\smalldisplayskips
\mathcal{R}^{\sigma} (f) = \frac{1}{\sigma} \int_{\mathcal{X} \times \mathcal{Y} } \phi \Big( \frac{y- f(\x)}{\sigma} \Big) \dif \rho(\x, y),
\end{equation*}
which can be viewed as a surrogate of the true modal regression risk $\mathcal{R}(f)$ since 
$\mathcal{R}(f) - \mathcal{R}^{\sigma} (f)  \rightarrow 0$ when $ \sigma \rightarrow 0$ \cite{feng2020learning}.

\subsection{Markovian Process}
Let $\{X_i\}_{i\geq 1}$ be a Markov chain on a general space $\mathcal{X}$ with invariant probability distribution $\pi$.
Let $P(x, \dif y)$ be a Markov transition kernel on a general space $(\mathcal{X}, \mathcal{B}(\mathcal{X}))$ and $P^*$ be its adjoint, i.e. $P^*(x, \dif y): = \frac{P(y, \dif x)}{\pi(\dif x)} \cdot \pi (\dif y)$. For a reversible Markov chain, $P$ is self-adjoint and coincides with $P^*$ and $(P+P^*)/2$. For a non-reversible Markov chain, $P$ is not self-adjoint, but $(P+P^*)/2$ is self-adjoint and associates with a reversible transition kernel \cite{rudolf2011explicit}.
Denote $\mathcal{L}_2(\pi)$ by the Hilbert space consisting of square integrable functions with respect to $\pi$. For any function $h: \mathcal{X} \rightarrow \mathbb{R}$, we write $\pi(h) :=\int h(x) \pi(\dif x)$. Define the norm of $h \in \mathcal{L}_{2}(\pi)$ as $\|h\|_{\pi} = \sqrt{ \langle h, h \rangle}$.
Let $P^t(x, \dif y ), (t \in \mathbb{N})$  be the $t$-step Markov transition kernel corresponding to $P$, then $P^t(x, S) = \mathrm{Pr}(X_{t+i} \in S| X_i = x)$ for $i \in \mathbb{N}, x \in \mathcal{X}$ and a measurable set $S$.

Following the above notations, we introduce the definitions of ergodicity and spectral gap for a Markov chain.

\begin{definition}
Let $M(x)$ be a  non-negative function. For an initial probability measure $\rho(\cdot)$ on $\mathcal{B}(\mathcal{X})$, a Markov chain is uniformly ergodic if 
\begin{equation}\label{eq_ergodic}
	\|P^{t}(\rho, \cdot) - \pi(\cdot)\|_{TV} \leq M(x) \rho^t
\end{equation}
for some $M(x) < \infty$ and $\rho< 1$, where $\|\cdot\|_{TV}$ denotes total variation norm.
\end{definition}
A Markov chain is geometrically ergodic if (\ref{eq_ergodic}) holds for some $t<1$, which eliminates the bounded assumption on $M(x)$. 

For a Markov chain with stationary distribution $\pi$, the spectrum of the chain is defined as
$\mathcal{S}:= \{ \bar{\lambda} \in \mathbb{C} \setminus 0:  (\bar{\lambda}I - P)^{-1} \textrm{ does not exist as
	a bounded linear operator on }\mathcal{L}_2(\pi) \}$.
For reversible chains, $\mathcal{S}$ lies on the real line.
\begin{definition}
(Spectral gap, absolute spectral gap and  pseudo spectral gap) \cite{paulin2015concentration}
The spectral gap for reversible chains is
\[ 
\gamma =
\begin{cases}
	1 - \sup \{\bar{\lambda}: \bar{\lambda} \in \mathcal{S}, \bar{\lambda} \neq 1 \}, &
	\begin{array}{l}
		\textrm{if eigenvalue 1 has}  \\
		\quad\textrm{multiplicity 1,}
	\end{array} \\
	0 & \textrm{ otherwise}
\end{cases}
\]
For both reversible and non-reversible chains, the absolute spectral gap is 
\[
\gamma_a = 
\begin{cases}
	1 - \sup \{|\bar{\lambda}|: \bar{\lambda} \in \mathcal{S}, \bar{\lambda} \neq 1 \}, & \begin{array}{l}
		\textrm{if eigenvalue 1 has}  \\
		\quad\textrm{multiplicity 1,}
	\end{array} \\
	0, &\textrm{ otherwise}
\end{cases}
\]


The pseudo spectral gap of a Markov operator $P$ is
\begin{equation*}
	\gamma_p := \max_{k \geq 1} \{ \gamma((P^*)^k P^k)/k \},
\end{equation*}  
where $ \gamma ((P^*)^k P^k)$ denotes the spectral gap of the self-adjoint operator  $(P^*)^k P^k$.
\end{definition}

\begin{remark}
The dependence of a Markov chain can be characterized by the spectral gap.  A small $\bar{\lambda}$ usually implies a fast convergence of the Markov chain towards its stationary distribution from a non-stationary initial distribution \cite{rudolf2011explicit}. Note that in the reversible case, $\gamma \geq \gamma_a$. The pseudo spectral gap is similar to the spectral gap in the sense that it allows to derive concentration bounds on MCMC empirical averages and is closely related to the mixing time  \cite{paulin2015concentration}. 
\end{remark}

\subsection{Regularized Modal Regression with Markov-Dependent Observations}
Define an integral operator $L_K: \mathcal{L}_2 \rightarrow \mathcal{L}_2$ associated with the kernel $K: \mathcal{X} \times \mathcal{X}  \rightarrow \mathbb{R}$ by
\begin{equation*}
L_K f(\x) = \int_{\mathcal{X}} K(\x, \cdot) f(\cdot) \dif \rho_X, ~ \x \in \mathcal{X}.
\end{equation*}
Suppose $\mathcal{X}$ is compact and $K$ is continuous, then $L_K L_K^{\top}$ is a self-adjoint positive operator with decreasing eigenvalues $\{\lambda_k^2\}_{k=1}^\infty$ with $\lambda_k \geq 0 $ and eigenfunctions $\{\psi_k\}_{k=1}^\infty$ forms an orthonormal basis of $\mathcal{L}_2$. With this setup, we further define $|L_K|^\beta = |L_K L_K^\top|^\frac{\beta}{2}$ with $|L_K|^\beta (\sum_{k=1}^{\infty} c_k \psi_k) = \sum_{k=1}^{\infty} c_k \lambda_k^\beta \psi_k $, $\{c_k\}_k \in \ell_2$.

Given samples  $\z $ and a continuous $K$,  the sample dependent hypothesis space (SDHS) is defined as
\begin{small}
	\begin{equation} \label{eq_SDHS}
		\mathcal{H}_{K,\z} = \Big\{  f = \sum_{i=1}^{m} \alpha_i K(\x_i, \cdot),  \aalpha = (\alpha_1, \alpha_2, \cdots, \alpha_m)^{\top} \in \mathbb{R}^m \Big\},
	\end{equation}
\end{small}
which has been extensively used in generalization analysis of regression and classification. SDHS does not require the kernel to be symmetric and semi-definite, hence  provides much flexibility and adaptivity for learning problems. It should be noted that the hypothesis 	$\mathcal{H}_{K,\z}$ can be expressed as the span of $K(\x, \cdot)$ over the inputs $\{\x_i\}_{i=1}^m$, which further implies that the hypothesis is determined by the coefficient $\alpha_i, i= 1, 2, \cdots, m$ once the kernel function is specified. Therefore,  regularized modal regression aims to solve the following optimization problem 
\begin{equation} \label{eq:RMR}
f_\z = \arg \max_{f \in \mathcal{H}_{K,\z}} \{ \mathcal{R}_\z^\sigma (f) -  \lambda \Omega_q (f)  \}, 
\end{equation}
where  $\lambda > 0 $ is a regularization parameter and $\Omega_q(f)$ is the coefficient regularizer, defined by 
\begin{equation*}
\Omega_q(f) = \inf\Big\{ \sum_{i=1}^{m} |\alpha_i|^q: f = \sum_{i=1}^{m} \alpha_i K_{\x_i} \subset \mathcal{H}_1  \Big\}
\end{equation*}
with $q = 1, 2$, where $\mathcal{H}_1$ is given in Definition \ref{def_banach}. Let $\K_i = (K(\x_1, \x_i), K(\x_2, \x_i), \cdots, K(\x_m, \x_i))$, then optimization model (\ref{eq:RMR}) can be reformulated as 
\begin{equation} \label{eq_coef}
		\aalpha^\z 
		= \mathop{\arg \max}_{ \aalpha  \in \mathbb{R}^m }  \Big\{  \frac{1}{m\sigma} \sum_{i=1}^{m} \phi \Big( \frac{y_i - \K_{i}^\top \aalpha }{\sigma} \Big) - \lambda \|\aalpha\|_q^q  \Big\}	
\end{equation}
with 
\begin{equation*}
f_\z =  \sum_{i=1}^{m} \alpha_{j}^\z K(\x_i, \cdot).
\end{equation*}

Note that  model  (\ref{eq_coef})  is reduced to a robust kernel machine to achieve sparseness when $q = 1$, which is a natural extension of  sparse kernel regression \cite{chen2018kernel,shi2019sparse}.  When $q = 2$, it is closely related to  kernel ridge regression by replacing modal regression criterion with the mean square error criterion.  In particular, when Gaussian kernel is employed for kernel density function, (\ref{eq_coef}) can be rewritten as  
\begin{equation*}
\aalpha^\z =  \mathop{\arg \max}_{ \aalpha  \in \mathbb{R}^m }  \Big\{  \frac{1}{m\sigma} \sum_{i=1}^{m} \exp  \Big\{ - \frac{ (y_i - \K_{i}^\top \aalpha)^2 }{\sigma} \Big\} - \lambda \|\aalpha\|_q^q  \Big\},
\end{equation*}
which is consistent with the sparse correntropy regression with coefficient-based regularization \cite{chen2018kernel}.  This problem can be solved efficiently through the Half Quadratic (HQ) \cite{nikolova2005analysis} optimization strategy.

\section{Theoretical Assessments } \label{sec:theory}
This section mainly concerns the theoretical property of regularized modal regression for Markov-dependent observations. Specifically, our goal is to bound the excess generalization error  $\mathcal{R}(f^*) - \mathcal{R}(f_\z)$ in the context of general Markov chain. To this end, we first introduce a Banach space $\mathcal{H}_1$, which contains all possible SDHS $\mathcal{H}_{K,\z}$ in  (\ref{eq_SDHS}).  

\begin{definition} \label{def_banach}
Define a Banach space $\mathcal{H}_1 =\big \{ f: f=\sum_{j=1}^{\infty} \alpha_j K(\x_j), \alpha_j \in \mathbb{R}, \{\x_j\} \subset X 
\big \}$ with the norm 
\begin{equation}
	\|f\| = \inf \Big\{  \sum_{j=1}^\infty  |\alpha_j|: f =\sum_{j=1}^\infty \alpha_j K_{\x_j}\Big\},
\end{equation}
\end{definition} 
It can be observed that $\mathcal{H}_1$ consists of continuous functions due  to the continuity of $K$. As an important measurement of capacity of a hypothesis space, covering number have been extensively studied in the work \cite{zhou2002cover,zhou2003capacity,steinwart2008support}. We adopt empirical covering number involved with $\mathcal{H}_1$ to get a tight bound for RMR estimator. 

\begin{definition}
(Empirical Covering Number \cite{wu2007multi})  Let $\mathcal{H}$ be a set of functions on $\mathcal{Z}$ and  samples $\z =  \{z_1, z_2, \cdots, z_m\} \subset  \mathcal{Z}$. The metric on $\mathcal{H}$ is denoted by 
\begin{equation*}
	d_{2, \z} (f, g) =  \Big\{  \frac{1}{m} \sum_{i=1}^{m} \big( f(z_i) - g(z_i) \big)^2  \Big\}^{1/2}  for  \quad f, g \in \mathcal{H}.
\end{equation*}
For any $ \varepsilon >0 $, the empirical covering number of $\mathcal{H}$ with respect to $d_{2, \z} (f, g)$ is 
\begin{equation*}
	\mathcal{N}_2(\mathcal{H}, \varepsilon) = \sup_{m \in \mathbb{N}} \sup_{\z} \mathcal{N}_{2, \z}(\mathcal{H}, \varepsilon) > 0,
\end{equation*}
where
\begin{equation*}
	\begin{split}
			\mathcal{N}_{2,\z}(\mathcal{H}, \varepsilon) : = \inf \Big\{l \in \mathbb{N}: \exists \{f_i\}_{i=1}^l \subset \mathcal{H} \ \textrm{such  that} \  \\
			\mathcal{H} = \bigcup_{i=1}^l  \big\{ f \in \mathcal{H}: d_{2,\z}(f, f_i) \leq \varepsilon \big\}  \Big\}.
	\end{split}
\end{equation*}
\end{definition}

Note that for any function set $\mathcal{H} \subset \mathcal{C}(\mathcal{X})$, the empirical covering number $\mathcal{N}_{2,\z}(\mathcal{H}, \varepsilon)$ can be bounded by $\mathcal{N}(\mathcal{H}, \varepsilon)$, the uniform covering number of $\mathcal{H}$ with the metric $\|\cdot\|_\infty$, due to the fact $ d_{2,\z}(f, g) \leq \|f - g\|_\infty$.  The function sets in our situation are balls of the SDHS in the form of $\mathcal{B}_R = \{f \in \mathcal{H}: \Omega_q^{\frac{1}{q}} (f)  \leq R \}$.


\begin{assumption} \label{assumption1}
(Complexity)	For any $\eta > 0$, there exists an exponent $s$  with $0 < s <2$ and $c_s >0 $ such that 
\begin{equation} \label{eq_complexity}
	\log \mathcal{N}_2(\mathcal{H}_{K,\z}, \eta) \leq c_s \eta^{-s}, ~~ \forall \eta > 0.
\end{equation}
\end{assumption}

\begin{assumption}  \label{assumption2}
(Non-zero spectral gap) The underlying Markov chain $\{X_i\}_{i=1}^n$ is stationary with unique invariant measure $\pi$ and admits an absolute spectral gap $1 - \gamma$.
\end{assumption}

\begin{assumption} \label{assumption3}
(Density) The conditional density of $\epsilon$ given $X$, i.e.  $p_{\epsilon | X}$ is second-order continuous differentiable and $\| p''_{\epsilon | X}\|_\infty$ is bounded.
\end{assumption}

\begin{assumption} \label{assumption4}
(Calibrated Modal Regression Kernel)	The representing function  $\phi$ satisfies: 1) $\forall  u \in \mathbb{R}, \phi(u) \leq \phi(0) < \infty$; 2)  $\phi $ is Lipschitz continuous with constant $L_\phi$; 3) $\int_\mathbb{R} \phi(u) \dif u = 1 $  with $\int_{\mathbb{R}} u^2 \phi(u) \dif u < \infty$.
\end{assumption}

Assumption \ref{assumption1} is a fairly standard assumption on describing the complexity of hypothesis space. It has been extensively studied in learning theory \cite{zhou2002cover,zhou2003capacity,cucker2001math}, from which we know for a $C^\infty$ kernel, (\ref{eq_complexity}) holds for any $s >0$.   Assumption \ref{assumption2} requires the underlying Markov chain admits an absolute spectral gap, which quantifies the converge speed of Markov chain towards its invariant distribution $\pi$. Assumption \ref{assumption3} is a general condition on  conditional density of $p''_{\epsilon|X}$ and conventional noise distributions satisfy this requirement.  Assumption \ref{assumption4} requires the represent  function to be bounded and Lipschitz continuous. Typical examples include the Gaussian kernel, Epanechnikov kernel, quadratic kernel and Triangular kernel.  The following comparison theorem \cite{feng2020learning} characterizes the relationship between excess modal risk and excess generalization risk. 

\begin{lemma} \label{lemma:comparison}
\cite{feng2020learning} Under assumption \ref{assumption3}, for any measurable function $f : \mathcal{X} \rightarrow \mathbb{R}$, it holds that
\begin{equation*}
	|\mathcal{R}(f^*) - \mathcal{R}(f) - (\mathcal{R}^\sigma(f^*) - \mathcal{R}^\sigma(f)) | \leq  C_1 \sigma^{2},
\end{equation*}
where $C_1 = \|p''_{\epsilon| X} \|_\infty \int_{\mathbb{R}} u^2 \phi(u) \dif u $.
\end{lemma}

A well-established approach for conducting error analysis of learning algorithms is error decomposition, where the generalization error is usually decomposed into sample error and approximation error. Considering the characteristic of SDHS,  we  formulate the error decomposition of RMR by introducing the stepping stone function$f_\lambda$, defined by
\begin{equation*}
f_\lambda = \arg \max_{f \in \mathcal{H}_{K,\z}} \{ \mathcal{R}^\sigma (f) -  \lambda \Omega_q(f)  \}, 
\end{equation*}
where $\lambda > 0$ is the regularization parameter.

\begin{proposition} \label{prop_decom}
Suppose $f_{\z}$ is produced by (\ref{eq:RMR}) based on Markov-dependent observations, and $f^* \in \mathcal{H}_{K,\z}$.Then 
\begin{equation*}
	\mathcal{R}(f^*) - \mathcal{R}(f_\z) \leq    \mathcal{S}_1(\z) + \mathcal{S}_2(\z) + C_1 \sigma^2  +  \lambda \Omega_q(f^*) ,
\end{equation*}
where $C_1 =  \|p''_{\epsilon| X} \|_\infty \int_{\mathbb{R}} u^2 \phi(u) \dif u$ and
\begin{equation*}
	\begin{split}
		\mathcal{S}_1(\z) &= \mathcal{R}_{\z}^\sigma (f^*) - \mathcal{R}_{\z}^\sigma (f_\lambda) - \{ \mathcal{R}^\sigma (f^*) - \mathcal{R}^\sigma (f_\lambda)   \}, \\
		\mathcal{S}_2(\z) &=  \mathcal{R}^\sigma (f^*) -  \mathcal{R}^\sigma (f_\z) - \{ \mathcal{R}_{\z}^\sigma (f^*) - \mathcal{R}_{\z}^\sigma (f_\z)   \}.
	\end{split}
\end{equation*}
\end{proposition}

With these settings, we now present theoretical results for RMR with Markov-dependent observations.
\begin{theorem} \label{th:1}
Let the Markov-dependent observations $\z$ be generated by (\ref{eq_model}) with invariant distribution $\pi$ and non-zero absolute spectral gap $\gamma_a >0$.  Suppose that \textbf{Assumptions} \ref{assumption1}-\ref{assumption4} are satisfied. Let $f^*$ lies in the range of  $L_K^\beta$ for some $\beta \in (0,2]$. Then for any $0 < \delta < 1$, the following inequality 
\begin{equation*}
	\begin{split}
			& \mathcal{R}^{\sigma}(f^*) -  \mathcal{R}^{\sigma}(f_\z) \leq  C \log(2/\delta)  \Big( (2\gamma_a - \gamma_a^2)^{-\frac{1}{2}} m^{-\frac{1}{2}} \sigma^{-\frac{1}{2}} \\
			& \qquad +  (2\gamma_a - \gamma_a^2)^{-\frac{1}{1+s}} m^{-\frac{1}{1+s}} \sigma^{- \frac{4+2s}{1+s}} R^{\frac{s+2}{s+1}}  + \sigma^2 +\lambda^{\frac{2\beta}{2+\beta}} \Big)
	\end{split}
\end{equation*}
holds with confidence at least $1 - \delta$, where  $C$ is a positive constant independent of $m, \sigma, \delta$.
\end{theorem} 
\begin{remark}
Theorem \ref{th:1} establishes the upper bound  for regularized modal regression in Markov-dependent setup. As far as we can tell, this is the first work in the literature.  It can be observed  that the corresponding generalization error relies on the spectral gap of underlying Markov chain, the capacity of hypothesis space, the regularization parameter $\lambda$ and  the bandwidth parameter $\sigma$. The dependence of the Markov chain is measured by a quantity $\gamma_a \in [0,1]$, denoting the norm of Markov operator (induced by transition kernel) acting on the $\mathcal{L}_2$ space with respect to the invariant distribution. It has been involved as constants in mean square error bound for Markov chain Monte Carlo  
\cite{rudolf2011explicit}, Hoeffding-type \cite{fan2018hoeffding} and Bernstein-type inequalities for Markov chains \cite{paulin2015concentration}. A non-zero spectral gap is closely related to other convergence criterion of Markov chains, e.g. geometrically ergodic, uniformly ergodic\cite{meyn2012markov}. Note that such a Markov chain can actually be generated by the so-called Markov sampling strategy \cite{gong2015learning,gong2020robust}, where a uniformly ergodic Markov chain can be generated from a given dataset without temporal relation. 
\end{remark}

\begin{theorem} \label{th:2}
Under the same conditions in Theorem \ref{th:1}, take $\theta =   \frac{2\beta}{8\beta + 5s\beta +2s +4} $,  $\lambda = (2\gamma_a - \gamma_a^2)^{-\frac{\theta}{\beta}} m^{-\frac{\theta}{\beta}}$ and $\sigma =(2\gamma_a - \gamma_a^2)^{-\frac{\theta}{2\beta}} m^{-\frac{\theta}{2\beta}} $. For any $ 0< \delta<1 $, the excess risk of RMR estimator $f_\z$ satisfies
\begin{equation*}
	\mathcal{R}(f^*) -  \mathcal{R}(f_\z) \leq 
	\hat{C}  \log(2/\delta) (2\gamma_a - \gamma_a^2) ^{- \frac{\theta}{\beta}}  m^{- \theta} 
\end{equation*}
with confidence at least $1 - \delta$, where $\hat{C}$ is a positive constant independent of $m, \sigma, \delta$.
\end{theorem} 
\begin{remark}
Theorem \ref{th:2} implies the estimation consistency of RMR when $\lambda, \sigma$ are properly specified. In particular, when $s \rightarrow 0$, $\beta =2 $ we see that the learning rate in Theorem \ref{th:2} is $\mathcal{O} \big( (2\gamma_a -\gamma_a^2)^{-\frac{1}{5}} m^{-\frac{1}{5}}  \big)$, which is faster than the result in \cite{wang2017regularized}, whose learning is $\mathcal{O} (m^{-\frac{1}{7}})$.  It is worth noting that the learning rate of RMR  in Markov-dependent samples would be discounted by a multiplicative coefficient  $ (2\gamma_a- \gamma_a^2)^{-\frac{1}{5}}$, which is determined by the convergence property of the underlying Markov chain. Generally, a small $\gamma$ will lead to a small coefficient, which means a Markov chain with fast converging speed has small generalization error. Note that the absolute spectral gap assumption can be relaxed to the pseudo spectral gap, the corresponding learning rate established in Theorem \ref{th:2} remains the same order but the multiplicative coefficient $2\gamma_a - \gamma_a^2$ is replaced by $\gamma_p$.
\end{remark}

\begin{remark}
It is well known that any bounded independent random variables $Z_i \in [a_i, b_i] ~ (a_i \leq b_i, a_i, b_i \in \mathbb{R} )$ can be seen as the transformations of  i.i.d. random variables $U_i \sim  \mathbf{Unif}[0,1]$ via the inverse cumulative distribution functions $F_{Z_i}^-1: [0, 1] \rightarrow [a_i, b_i]$, i.e. $Z_i  = F_{Z_i}^{-1}(U_i)$. Hence, the i.i.d. sequence $\{U_i \}_{i \geq 1}$ can be regarded as a stationary Markov chain on the state space $[0,1]$ with invariant measure $ \pi(dy) = dy$ and transition kernel $P(x, dy) = dy$. This Markov chain  has  $\gamma  = 1$. In this case,  the generalization error  in Theorem \ref{th:2} reduces to the classical  i.i.d. case, i.e. $\mathcal{O}(m^{-\frac{1}{5}})$. Note that such a learning rate is still better than the result in \cite{wang2017regularized}. The main reason is that we use empirical covering number to carefully characterize the capacity of function space while Wang et al. \cite{wang2017regularized} adopts the Rademacher complexity as the measurement. Some regularity conditions can be imposed on the kernel function to further improve the learning rate.
\end{remark}

To evaluate the robustness of RMR within Markov-dependent observations, we introduce the concept of breakdown point \cite{donoho1982breakdown}, which measures the proportion of bad data in a dataset that an 
estimator can tolerate before returning arbitrary value. 

Given a sample set $\z = \{(\x_i, y_i)\}_{i=1}^m$, the corrupted sample set $\z \cup \z'$ is constructed by adding $n$ arbitrary points $\z' = \{(\x_{m+j}, y_{m+j})\}_{j=1}^n $, which contain a fraction $\frac{n}{m+n}$ of bad values. The finite sample contamination breakdown point $\epsilon(\aalpha_\z)$ is defined as 
\begin{equation}
	\varepsilon^*(\aalpha_\z)  = \mathop{\min_{1 \leq n \leq m}} \Big\{\frac{n}{m+n}: \sup_{\z'} \|\aalpha_{\z \cup \z'}\|_2 = \infty \Big\}, 
\end{equation}
\begin{theorem} \label{th:3}
Suppose $\phi(u) = \phi(-u)$ and $\phi(t) \rightarrow 0$ when $|t| \rightarrow \infty$. For a given Markov-dependent observations $\z$, and $\lambda, \sigma$, let 
\begin{equation*}
	N = \phi(0)^{-1} \sum_{i=1}^{m} \phi \Big(   \frac{y_i - \K_i^{\top} \aalpha_\z }{\sigma}\Big) - \lambda  \phi(0)^{-1} m \sigma\|\aalpha_\z\|_q^q.
\end{equation*}
Then the finite sample contamination breakdown point of $\aalpha_\z$ in (\ref{eq_coef}) is 
\begin{equation*}
	 \varepsilon^*(\aalpha_\z) = \frac{n^*}{m+n^*},
\end{equation*}
where $n^*$ is an integer satisfying $\lceil N \rceil \leq n^* \leq  \lfloor N \rfloor +1$, $\lceil a \rceil$ denotes the largest integer not greater than $a$ and $\lfloor a \rfloor$ denotes the smallest integer not less than $a$.  
\end{theorem}
\begin{remark}
	Theorem \ref{th:3} indicates that the breakdown point relies on $\phi(\cdot)$, the turning parameter $\lambda,\sigma$ and the sample configuration. As pointed out in \cite{huber1992robust}, the breakdown point can be quite high if the bandwidth parameter is only determined by training samples. However, with appropriate choice of $\lambda$ and $\sigma$ through some data driven strategies, RMR can still achieve a satisfactory learning rate and robustness. 
\end{remark}

\section{Proofs} \label{sec:proofs}
This section presents the proof details of the main theorems.  To be clear, we first list several useful lemmas which will be used in the proofs. 


\begin{lemma} \cite{paulin2015concentration}  \label{Lemma:Markov_Bernstein}
(Bernstein inequality for reversible Markov Chains)	Let $X_1, X_2, \cdots, X_m$ be a stationary reversible Markov chain with invariant distribution $\pi$ and absolute spectral gap $\gamma_a$. Suppose that $f_1, f_2, \cdots, f_m \in \mathcal{L}_2(\pi)$ with $|f_i -  \mathbb{E}_\pi(f_i) | \leq C $, denote $S := \sum_{i=1}^m f_i(X_i)$ and $V_S := \sum_{i=1}^{m} \textrm{Var}_{\pi}(f_i)$, then for any $t > 0$,
\begin{equation}
	\mathbb{P}_\pi \Big( |S  - \mathbb{E}_\pi (S) | \geq t  \Big) \leq 2 \exp \Big( -\frac{t^2(2\gamma_a- \gamma_a^2)}{8V_S + 20 C t} \Big).
\end{equation}
\end{lemma}


\begin{lemma} \cite{cucker2002} \label{lemma:unique}
Let $c_{1}, c_{2} > 0$, and $p_{1} > p_{2} >0$. Then, the equation $x^{p_{1}} - c_{1}x^{p_{2}} - c_{2} = 0$ has unique positive zero $x^{*}$. In addition $x^{*} \leq \max\{ (2c_{1})^{1/(p_{1} - p_{2})}, (2c_{2})^{1/p_{1}}\}$.
\end{lemma}

\subsection{Proof of Theorem \ref{th:1}}
\begin{proof}
The proof of Theorem \ref{th:1} consists of three steps below.

\textbf{Step I:	}  Bounding  $\mathcal{S}_1(\z)$.
Define a random variable 
\begin{equation*}
	\xi_1  := \sigma^{-1} \phi\Big( \frac{y - f^*(\x)}{\sigma}  \Big) - \sigma^{-1} \phi\Big(\frac{y - f_{\lambda}(\x)}{\sigma}   \Big), \z \in \mathcal{Z}.
\end{equation*}
According to the boundedness assumption of $\phi$, it is easy to check that $|\xi_1(\z)| \leq 2 \|\phi\|_\infty /\sigma$. Furthermore, we see that
\begin{equation*}
	\begin{split}
		\textrm{Var}(\xi_1) &= \mathbb{E} \Big[ \sigma^{-1} \phi\Big( \frac{y - f^*(\x)}{\sigma}  \Big) - \sigma^{-1} \phi\Big(\frac{y - f_{\lambda}(\x)}{\sigma}  \Big) \Big]^2 \\
		& \leq  2 \frac{\|\phi\|_\infty}{\sigma} (\mathcal{R}^\sigma(f^*)  + \mathcal{R}^\sigma(f_\lambda)).
	\end{split}
\end{equation*}
By Theorem 9 in \cite{feng2020learning}, we have
\begin{equation*}
	|\mathcal{R}^\sigma(f) - \mathcal{R}(f) |  \leq \frac{C_1 \sigma^2}{2},
\end{equation*}
which implies $\mathcal{R}^\sigma(f^*) \leq \mathcal{R}(f^*) +  \frac{C_1}{2} \sigma^2$
and $\mathcal{R}^\sigma(f_\lambda) \leq \mathcal{R}(f_\lambda) +  \frac{C_1}{2} \sigma^2$, 
where $C_1$ is given in Lemma \ref{lemma:comparison}. These two inequalities together with the fact $\sigma \leq 1$ yield
\begin{equation*}
	\begin{split}
		\textrm{Var}(\xi_1) & \leq   \frac{2 \|\phi\|_\infty}{\sigma} (\mathcal{R}(f^*)+ \mathcal{R}(f_\lambda) + C_1 \sigma^2) \\
		&\leq    \frac{2 \|\phi\|_\infty}{\sigma} (p_{f^*}(0) + p_{f_\lambda}(0) + C_1 \sigma^2) \\
		& \leq C_2 \sigma^{-1},
	\end{split}
\end{equation*}
where $C_2 = 2\|\phi\|_\infty (p_{f^*}(0) + p_{f_\lambda}(0) + C_1)$. Now applying Lemma \ref{Lemma:Markov_Bernstein} to the random variable $\xi_1$, we have
\begin{equation*}
	\mathcal{S}_1   \leq \frac{20 \|\phi\|_\infty  \log(2/\delta)}{m \sigma(2\gamma_a - \gamma_a^2)} +   2\sqrt{\frac{ 2C_2 \log(2/\delta)}{m \sigma(2 \gamma_a- \gamma_a^2)}  }.
\end{equation*}
with confidence at least  $1 - \delta$.

\textbf{Step II:} Bounding $\mathcal{S}_2(\z)$. To this end, we first prove that under assumptions \ref{assumption1} and \ref{assumption2}, for  any $f \in \mathcal{B}_R$ with $R \geq 1$ and $\varepsilon \geq C_1 \sigma^2$,  with confidence at least  $1 - \delta$, it holds
\begin{small}
\begin{equation} \label{eq_cover}
	\begin{split}
		&\mathbb{P}_{\z \in \mathcal{Z}^m} \Big\{ 	\frac{\mathcal{R}^{\sigma}(f^*) -  \mathcal{R}^{\sigma}(f) - (\mathcal{R}_\z^{\sigma}(f^*) -  \mathcal{R}_\z^{\sigma}(f))}{ \sqrt{\mathcal{R}^{\sigma}(f^*) -  \mathcal{R}^{\sigma}(f) + 2 \varepsilon}}   > 4\sqrt{\varepsilon} \Big\} \\
		\leq ~&\mathcal{N}_2 \left(\mathcal{B}_R, r \right)   \exp\Big\{- \frac{ (2\gamma_a - \gamma_a^2)m \varepsilon }{40(M+1)^2R^2 ( L_\phi \sigma^{-4} +  L_\phi \sigma^{-2})} \Big\},
	\end{split}
\end{equation}
\end{small}
where $r  = \frac{\sigma^2 \varepsilon}{L_\phi (M+1)R}$. 
To this end, we  introduce a random variable defined by	
\begin{equation*}
	\xi_2  = \sigma^{-1} \phi \Big( \frac{y - f^*(\x)}{\sigma} \Big) -  \sigma^{-1} \phi \Big( \frac{y - f(\x)}{\sigma} \Big),
\end{equation*}
then it is easy to verify that  $\mathbb{E} \xi_2 = \mathcal{R}^{\sigma}(f^*) - \mathcal{R}^{\sigma}(f)$ and for $\forall f \in \mathcal{B}_R$, $|\xi_2 |   \leq \frac{L_\phi}{\sigma^2}  \|f ^*- f\|_\infty \leq \frac{L_\phi}{\sigma^2} (M+1)R$, $|\xi_2 - \mathbb{E}\xi_2|  \leq \frac{2 L_\phi}{\sigma^2} (M+1)R $ and $\textrm{Var}(\xi_2) \leq \mathbb{E} \xi_2^2 \leq \frac{L_\phi^2}{\sigma^4} \|f ^*- f\|_\infty ^2$.
Let $\{f_j\}_{j=1}^J$ be an $r$-net  of the set $\mathcal{B}_R$ with $J$ being the covering number of $\mathcal{N}_2(\mathcal{B}_R, r)$, and define 
\begin{equation*}
	\mu = \sqrt {\mathcal{R}^{\sigma}(f^*) -  \mathcal{R}^{\sigma}(f_j) + 2 \varepsilon }.
\end{equation*}
According to  Lemma \ref{Lemma:Markov_Bernstein}, we get the following conclusion
\begin{small}
	\begin{align*}	
		&\mathbb{P}_{\z \in \mathcal{Z}^m} \Big\{ \frac{\mathcal{R}^{\sigma}(f^*) -  \mathcal{R}^{\sigma}(f_j) - (\mathcal{R}_\z^{\sigma}(f^*) -  \mathcal{R}_\z^{\sigma}(f_j))}{ \sqrt{\mathcal{R}^{\sigma}(f^*) -  \mathcal{R}^{\sigma}(f_j) + 2 \varepsilon}}   > \sqrt{\varepsilon}  \Big\} \\
		\leq  &\exp \Big\{  -\frac{ (2\gamma_a - \gamma_a^2) m \mu^2 \varepsilon}{8 L_\phi^2\sigma^{-4}  \|f^* - f_j\|_\infty^2 + 40 L_\phi \sigma^{-2}(M+1)R \mu \sqrt{\varepsilon} } \Big\}	 \\
		\leq & \exp\! \Big\{\! -\!\frac{ (2\gamma_a - \gamma_a^2)  m \mu^2 \varepsilon}{ 8 L_\phi^2\sigma^{-4} (M\!+\!1)^2 R^2 \mu^2 \!+\! 40 L_\phi \sigma^{-2}(M\!+\!1)^2 R^2 \mu \sqrt{\varepsilon} } \!\Big\}	\\
		\leq & \exp \Big\{- \frac{ (2\gamma_a - \gamma_a^2)  m \varepsilon }{40(M+1)^2 R^2 ( L_\phi^2 \sigma^{-4} + L_\phi \sigma^{-2})}\Big\}.
	\end{align*}
\end{small}
Since 
\begin{equation*}
	\mu^2 = \mathcal{R}^{\sigma}(f^*) -  \mathcal{R}^{\sigma}(f_j) + 2 \varepsilon >\mathcal{R}^{\sigma}(f^*) -  \mathcal{R}^{\sigma}(f_j) +  \varepsilon \geq \varepsilon,
\end{equation*}
there exists some $j$ such that $\|f - f_j\|\infty \leq \frac{\sigma^2 \varepsilon}{L_\phi (M+1)R}$ for any $f \in \mathcal{B}_R$, hence both $|\mathcal{R}^\sigma(f) - \mathcal{R}^\sigma(f_j) |$ and $|\mathcal{R}_\z^\sigma(f) - \mathcal{R}_\z^\sigma(f_j)|$ can be bounded by $\varepsilon$, then we have the following inequalities
\begin{equation*}
	\begin{split}
		\frac{|\mathcal{R}_\z^{\sigma}(f^*) - \mathcal{R}_\z^{\sigma}(f) - (\mathcal{R}_\z^{\sigma}(f^*) - \mathcal{R}_\z^{\sigma}(f_j) )|}{\sqrt{\mathcal{R}^{\sigma}(f^*)- \mathcal{R}^{\sigma}(f) + 2\varepsilon} } & \leq \sqrt{\varepsilon},	\\
		\frac{|\mathcal{R}^{\sigma}(f^*) - \mathcal{R}^{\sigma}(f) - (\mathcal{R}^{\sigma}(f^*) - \mathcal{R}^{\sigma}(f_j) )|}{\sqrt{\mathcal{R}^{\sigma}(f^*)- \mathcal{R}^{\sigma}(f) + 2\varepsilon } } & \leq \sqrt{\varepsilon}.
	\end{split}
\end{equation*}
These  two inequalities together with the fact $\varepsilon < \mathcal{R}^{\sigma}(f^*) -  \mathcal{R}^{\sigma}(f) + 2 \varepsilon$ yield the following inequality
\begin{equation*}
	\begin{split}
	&	\mathcal{R}^{\sigma}(f^*) -  \mathcal{R}^{\sigma}(f_j) + 2 \varepsilon = \mathcal{R}^{\sigma}(f^*) -   \mathcal{R}^{\sigma}(f_j)  - \\
	 &\qquad \qquad \qquad (\mathcal{R}^\sigma(f^*) - \mathcal{R}^\sigma(f))
		+ \mathcal{R}^{\sigma}(f^*) \!- \! \mathcal{R}^{\sigma}(f) \!+\! 2 \varepsilon \\
		& \leq \sqrt{\varepsilon} \sqrt{ \mathcal{R}^{\sigma}(f^*) -  \mathcal{R}^{\sigma}(f) + 2 \varepsilon} +  \mathcal{R}^{\sigma}(f^*) -  \mathcal{R}^{\sigma}(f) + 2 \varepsilon\\
		& \leq 2 ( \mathcal{R}^{\sigma}(f^*) -  \mathcal{R}^{\sigma}(f) + 2 \varepsilon),
	\end{split} 
\end{equation*}
hence
\begin{equation*}
	\frac{\mathcal{R}^{\sigma}(f^*) -  \mathcal{R}^{\sigma}(f) - (\mathcal{R}_\z^{\sigma}(f^*) -  \mathcal{R}_\z^{\sigma}(f))}{ \sqrt{\mathcal{R}^{\sigma}(f^*) -  \mathcal{R}^{\sigma}(f) + 2 \varepsilon}}   > 4\sqrt{\varepsilon}
	\end{equation*}
for $\forall f \in \mathcal{B}_R$. We further get 
\begin{equation*}
	\frac{\mathcal{R}^{\sigma}(f^*) -  \mathcal{R}^{\sigma}(f_j) - (\mathcal{R}_\z^{\sigma}(f^*) -  \mathcal{R}_\z^{\sigma}(f_j))}{ \sqrt{\mathcal{R}^{\sigma}(f^*) -  \mathcal{R}^{\sigma}(f_j) + 2 \varepsilon}}   > \sqrt{\varepsilon},
\end{equation*}
which implies 
\begin{small}
\begin{equation*}
	\begin{split}
		&\mathbb{P}_{\z \in \mathcal{Z}^m} \Big\{ 	\frac{\mathcal{R}^{\sigma}(f^*) -  \mathcal{R}^{\sigma}(f) - (\mathcal{R}_\z^{\sigma}(f^*) -  \mathcal{R}_\z^{\sigma}(f))}{ \sqrt{\mathcal{R}^{\sigma}(f^*) -  \mathcal{R}^{\sigma}(f) + 2 \varepsilon}}   > 4\sqrt{\varepsilon}  \Big\} \\
		\leq &\sum_{i=1}^J 	\mathbb{P}_{\z \in \mathcal{Z}^m} \!\Big\{\! \frac{\mathcal{R}^{\sigma}(f^*) \!-\!  \mathcal{R}^{\sigma}(f_j) \!-\! (\mathcal{R}_\z^{\sigma}(f^*) \!-\!  \mathcal{R}_\z^{\sigma}(f_j))}{ \sqrt{\mathcal{R}^{\sigma}(f^*) -  \mathcal{R}^{\sigma}(f_j) + 2 \varepsilon}}   > \sqrt{\varepsilon}\Big \}  \\
		\leq &\mathcal{N}_2 \left(\mathcal{B}_R, r \right)   \exp\Big\{- \frac{ (2\gamma_a - \gamma_a^2)m \varepsilon }{40(M+1)^2 R^2 ( L_\phi^2 \sigma^{-4} + L_\phi \sigma^{-2})} \Big\}.
	\end{split}
\end{equation*}
\end{small}
We know from (\ref{eq_cover}) that 
\begin{small}
\begin{equation} \label{eq_uniform}
	\begin{split}
		& \mathbb{P}_{\z \in \mathcal{Z}^m} \! \Big\{ \! \sup_{f_\z  \in \mathcal{B}_R} \!
		\frac{\mathcal{R}^{\sigma}(f^*) \!-\!  \mathcal{R}^{\sigma}(f_\z) \!-\! (\mathcal{R}_\z^{\sigma}(f^*) \!-\!  \mathcal{R}_\z^{\sigma}(f_\z))} { \sqrt{\mathcal{R}^{\sigma}(f^*) \!-\!  \mathcal{R}^{\sigma}(f_\z) \!+\! 2 \varepsilon}} \!>\! 4 \sqrt{\varepsilon} \!\Big\} \\
		&\leq \mathcal{N}_2 \left(\mathcal{B}_R, r \right)   \exp \Big\{- \frac{ (2\gamma_a -\gamma_a^2) m \varepsilon }{40(M+1)^2 R^2 ( L_\phi^2 \sigma^{-4} +  L_\phi \sigma^{-2})} \Big\} \\
		&\leq  \exp \Big\{  c_s \Big(\frac{1}{r} \Big)^s   - \frac{ (2\gamma_a -\gamma_a^2)m \varepsilon }{40(M+1)^2 R^2 ( L_\phi^2 \sigma^{-4} + L_\phi \sigma^{-2})}\Big\},
	\end{split}
\end{equation}
\end{small}
Set  the last term of inequality (\ref{eq_uniform}) equal to $\delta$, and we get

\begin{align*}
	\varepsilon^{s+1}  & - \frac{40(M+1)^2 R^2 ( L_\phi^2 \sigma^{-4} +  L_\phi \sigma^{-2}) \log(2/\delta)}{  (2\gamma_a- \gamma_a^2)m} \cdot \varepsilon^s  \\ 
	& -  \frac{40 c_s L_\phi^{s+2} ( \sigma^{-4-2s} \!+\!  \sigma^{-2-2s}) (M\!+\!1 )^{s+2}  R^{s+2} }{ (2\gamma_a - \gamma_a^2)m} \!=\! 0.
\end{align*}
By Lemma \ref{lemma:unique}, we obtain the upper bound of the smallest positive solution $\varepsilon^\Delta$ for the above equation, i.e.
\begin{equation*}
	\varepsilon^\Delta := C_3 (2\gamma_a- \gamma_a^2)^{-\frac{1}{1+s}} \sigma^{- \frac{4+2s}{1+s}} m^{-\frac{1}{1+s}}R^{\frac{s+2}{s+1}} \log(2/\delta),
\end{equation*}
where $C_3 := \max  \big\{ 80 (M+1) L_\phi,  (80c_s)^{\frac{1}{1+s}} (M+1)^{\frac{s+2}{s+1}} L_\phi^{\frac{s+2}{s+1}}  \big \}$. Then, we have
for $f_\z \in \mathcal{B}_R$
\begin{equation} \label{eq_S2}
	\begin{split}
		\mathcal{S}_2(\z) & = \mathcal{R}^{\sigma}(f^*) -  \mathcal{R}^{\sigma}(f_\z) - (\mathcal{R}_\z^{\sigma}(f^*) -  \mathcal{R}_\z^{\sigma}(f_\z)) \\
		&\leq 4 \sqrt{\varepsilon^\Delta} \cdot \sqrt{\mathcal{R}^{\sigma}(f^*) -  \mathcal{R}^{\sigma}(f_\z) + 2 \varepsilon^\Delta} \\
		&\leq  \frac{1}{2} (\mathcal{R}^{\sigma}(f^*) -  \mathcal{R}^{\sigma}(f_\z) ) + 9 \varepsilon^\Delta.
	\end{split}
\end{equation}

\textbf{Step III:} 	Let $\{\psi_1, \psi_2, \cdots\}$ be an orthonormal basis of $\mathcal{L}_{\rho_X}^2(X)$ and $\{\lambda_1, \lambda_2, \cdots\}$ be the corresponding eigenvalue with descending order.  Recall that $f^* \in L_K^\beta g$ for some $ 0<\beta \leq 2 $ and $g \in \mathcal{L}_{\rho_X}^2$, then $f^* = \sum_{\lambda_k \geq 0} \alpha_k \lambda_k^\beta \psi_k$ and  $g$ can be uniquely written as $g = \sum_{\lambda_k \geq 0} \alpha_k \psi_k $ with $\|g\|_{\mathcal{L}_{\rho_X}^2}^q = \sum_{\lambda_k \geq 0} |\alpha_k|^q  \leq \infty$. Assume that $ \lambda_1 \leq \lambda^{\frac{\beta-2}{2+\beta}} $, we get $\Omega_q(f^*) \leq \sum_{\lambda_k \geq 0} |\alpha_k|^q \lambda_k^\beta   \leq  \|g\|_{\mathcal{L}_{\rho_X}^2}^q \lambda^{\frac{\beta-2}{2+\beta}}$. 
Based on the estimations in \textbf{Step I} and \textbf{II}, we see that 
with confidence at least $1 -  \delta$, 
\begin{equation*}
	\begin{split}
		&\mathcal{R}(f^*) -  \mathcal{R}(f_\z) \leq  \frac{20 \|\phi\|_\infty  \log(2/\delta)}{m \sigma(2\gamma_a- \gamma_a^2)} +   2\sqrt{\frac{ 2C_2 \log(2/\delta)}{m \sigma(2\gamma_a- \gamma_a^2)} } \\
		& \qquad + 18C_3 (2\gamma_a- \gamma_a^2)^{-\frac{1}{1+s}} \sigma^{- \frac{4+2s}{1+s}} m^{-\frac{1}{1+s}}R^{\frac{s+2}{s+1}} \log(2/\delta) \\
		&\qquad + 2\lambda^{\frac{2\beta}{2+\beta}} +2 C_1 \sigma^2 \\
		&\leq C \log(2/\delta)  \Big( (2\gamma_a - \gamma_a^2)^{-\frac{1}{2}} m^{-\frac{1}{2}} \sigma^{-\frac{1}{2}} +  (2\gamma_a - \gamma_a^2)^{-\frac{1}{1+s}} \\
		 &\qquad \cdot m^{-\frac{1}{1+s}} \sigma^{- \frac{4+2s}{1+s}} R^{\frac{s+2}{s+1}}  + \sigma^2 +\lambda^{\frac{2\beta}{2+\beta}} \Big),
	\end{split}
\end{equation*}
where $C \!=\! 180C_1 C_3 M^2 \|\phi\|_\infty \|g\|_{\mathcal{L}_{\rho_X}^2}^q $ is a constant independent of $m, \delta, \sigma $ and $ \lambda $. We complete the proof.
\end{proof}

\subsection{Proof of Theorem \ref{th:2}}
\begin{proof}
From the definition of $f_\z$, we know that $\mathcal{R}_\z^\sigma (f_\z) - \lambda \Omega_q(f_\z) \geq \mathcal{R}_\z^\sigma(0)$,
then  $\lambda  \Omega_q(f_\z)  \leq 	\mathcal{R}_\z^\sigma (f_\z) - 	\mathcal{R}_\z^\sigma (0) \leq \frac{2 \|\phi\|_\infty}{\sigma},$  which implies
$\Omega_q(f_\z) \leq  2\|\phi\|_\infty \lambda^{-1} \sigma^{-1}$. Hence  taking $R =  2\|\phi\|_\infty \lambda^{-1} \sigma^{-1}$ together with Theorem \ref{th:1} yield 
\begin{align*}
	&\mathcal{R}(f^*) -  \mathcal{R}(f_\z) \leq 
	\hat{C} \log(2/\delta) \Big(  (2\gamma_a-\gamma_a^2)^{-\frac{1}{2}}m^{-\frac{1}{2}} \sigma^{-\frac{1}{2}} \\
	&+ (2\gamma_a -\gamma_a^2)^{-\frac{1}{1+s}}  m^{-\frac{1}{1+s}}  \sigma^{-\frac{6+3s}{1+s}} \lambda^{-\frac{s+2}{s+1}}  + \sigma^2 +\lambda^{\frac{2\beta}{\beta+2}}  \Big).	
\end{align*}
By taking $\lambda = (2\gamma_a - \gamma_a^2)^{-\frac{\theta}{\beta}} m^{-\frac{\theta}{\beta}}$,  $\sigma =(2\gamma_a - \gamma_a^2)^{-\frac{\theta}{2\beta}} m^{-\frac{\theta}{2\beta}} $ and 	
$\theta =   \frac{2\beta}{8\beta + 5s\beta +2s +4} $
with confidence at least $1 - \delta$, it holds
\begin{equation*}
	\mathcal{R}(f^*) -  \mathcal{R}(f_\z) \leq 
	\hat{C}  \log(2/\delta) (2\gamma_a - \gamma_a^2) ^{-\frac{\theta}{\beta}}  m^{-\theta} . 
\end{equation*}
This completes the proof.
\end{proof}

\subsection{Proof of Theorem \ref{th:3}}
\begin{proof}
Observe that the RMR optimization problem (\ref{eq_coef}) is equivalent to 
\begin{equation}
	\max_{ \aalpha  \in \mathbb{R}^m } \Big\{   \phi(0)^{-1} \sum_{i=1}^{m} \phi \Big(   \frac{y_i - \K_i^{\top} \aalpha_\z }{\sigma}\Big) - \lambda  \phi(0)^{-1} m \sigma\|\aalpha_\z\|_q^q \Big\}.
\end{equation}
Let $\phi^*(t) = \phi(t) / \phi(0)$. Then for any $t$, $\phi^*(t) \leq \phi^*(0) = 1$ and $\phi^*(\cdot)$ decreases monotonely toward both sides and that $\phi^*(t) = 0$ for $|t| \rightarrow \infty$.  

We first show that $\aalpha_{\z \cup \z'}$ is bounded when $n < N$. To this end, suppose there exists a $\nu > 0$ such that $ n + m\nu < N$.  Let $\phi^*(t) \leq \nu$ for $t \geq C$ and $\aalpha$ be any real vector such that $ |y -  \K^\top \aalpha| \geq C $ for all $ (\x, y) \in \z$. Then we have
\begin{equation} \label{eq:b1}
	\sum_{i=1}^{m+n} \phi^*(y_i - \K_i^{\top} \aalpha_\z) - \lambda m \sigma \|\aalpha_\z\|_q^q \geq N,
\end{equation}
and 
\begin{equation} \label{eq:b2}
	\begin{split}
			&\sum_{i=1}^{m+n} \phi^*(y_i - \K_i^{\top} \aalpha) - \lambda m \sigma \|\aalpha\|_q^q \\ 
		&\leq  \sum_{i=m+1}^{m+n}\phi^*(y_i - \K_i^{\top} \aalpha ) + \sum_{i=1}^{m} \phi^*(y_i - \K_i^\top \aalpha)  \\  
		& \leq n + m\nu.
	\end{split}
\end{equation} 
From (\ref{eq:b1}) and (\ref{eq:b2}), we know that $\aalpha_{\z \cup \z'}$ must satisfies $| y - \K^\top \aalpha_{\z \cup \z'} | < C$ for a sample in $\z$.

On the other hand, if $n > N$, let $\nu > 0 $,  such that $ n - n \nu > N$, and let $C$ be such that $\phi^*(t) \leq \nu$ for $| t| \geq C$. Assume that all points in $\z'$ are the same and satisfy $y = \K_i^{\top} \aalpha^*$. Let $\aalpha $ be any vector such that $| y_{m+1} - \K_{m+1}^\top \aalpha | < C$. Then $\sum_{i=1}^{m+n} \phi^*(y_i -  \K_{i}^\top \aalpha)  -  \lambda m \sigma \|\aalpha\|_q^q \leq N + n \nu$ and $\sum_{i=1}^{m+n} \phi^*(y_i -  \K_{i}^\top \aalpha^*) - \lambda m \sigma \|\aalpha^*\|_q^q \geq n$.
These inequalities imply that $|y_{m+1} -  \K_{m+1}^\top \aalpha_{\z \cup \z'} | \leq C$. Hence $\aalpha_{\z \cup \z'}$ is bounded as $n < N$. Observe that $\|\aalpha_{\z \cup \z'}\|  \rightarrow \infty$  when $ y_{m+1} \rightarrow \infty$ with $\K_{m+1}$ fixed, and we have the breakdown.
\end{proof}

\section{Conclusions} \label{sec:conclude}
In this paper, we investigate the generalization performance of regularized modal regression under Markov-dependence setup.  The statistical consistency is established and an explicit learning rate is given as well.  Our results show that the Markov dependence impacts on the generalization error in the way that sample size would be discounted by a multiplicative factor depending on the spectral gap of underlying Markov chain. Moreover, the study brings us some insights into robust regression within Markov-dependent setup. It will be interesting to improve the learning rate obtained in current study by imposing some regularity conditions on the kernel function.

\section{Acknowledgments}

This work was supported by National Key Research and Development Program of China(2020AAA0108800), National Natural Science Foundation of China (62106191, 12071166, 61772409, 62050194, 61721002), Innovation Research Team of Ministry of Education (IRT\_17R86), Project of China Knowledge Centre for Engineering Science and Technology and Project of Chinese academy of engineering ``The Online and Offline Mixed Educational ServiceSystem for ‘The Belt and Road’ Training in MOOC China’’.

\bibliographystyle{aaai}
\bibliography{MCC}	

\begin{thebibliography}{33}
\providecommand{\natexlab}[1]{#1}

\bibitem[{Chen and Wang(2018)}]{chen2018kernel}
Chen, H.; and Wang, Y. 2018.
\newblock Kernel-based sparse regression with the correntropy-induced loss.
\newblock \emph{Applied and Computational Harmonic Analysis}, 44(1): 144--164.

\bibitem[{Chen et~al.(2016)Chen, Genovese, Tibshirani, Wasserman
  et~al.}]{chen2016nonparametric}
Chen, Y.-C.; Genovese, C.~R.; Tibshirani, R.~J.; Wasserman, L.; et~al. 2016.
\newblock Nonparametric modal regression.
\newblock \emph{The Annals of Statistics}, 44(2): 489--514.

\bibitem[{Cucker and Smale(2001)}]{cucker2001math}
Cucker, F.; and Smale, S. 2001.
\newblock On the mathematical foundations of learning.
\newblock \emph{Bull. Amer. Math. Soc.}, 39(1): 1--49.

\bibitem[{Cucker and Smale(2002)}]{cucker2002}
Cucker, F.; and Smale, S. 2002.
\newblock Best choices for regularization parameters in learning theory: on the
  bias-variance problem.
\newblock \emph{Found. Comput. Math.}, 2(4): 413--428.

\bibitem[{Donoho(1982)}]{donoho1982breakdown}
Donoho, D.~L. 1982.
\newblock Breakdown properties of multivariate location estimators.
\newblock Technical report, Ph. D. Qualifying paper, Department of Statistics,
  Harvard University.

\bibitem[{Fan, Jiang, and Sun(2018)}]{fan2018hoeffding}
Fan, J.; Jiang, B.; and Sun, Q. 2018.
\newblock Hoeffding's lemma for Markov Chains and its applications to
  statistical learning.
\newblock \emph{arXiv preprint arXiv:1802.00211}.

\bibitem[{Feng, Fan, and Suykens(2020)}]{feng2020learning}
Feng, Y.; Fan, J.; and Suykens, J. 2020.
\newblock A statistical learning approach to modal regression.
\newblock \emph{J. Mach. Learn. Res.}, 21: 1--35.

\bibitem[{Gong, Xi, and Xu(2020)}]{gong2020robust}
Gong, T.; Xi, Q.; and Xu, C. 2020.
\newblock Robust Gradient-Based Markov Subsampling.
\newblock In \emph{Proceedings of the AAAI Conference on Artificial
  Intelligence}, volume~34, 4004--4011.

\bibitem[{Gong, Zou, and Xu(2015)}]{gong2015learning}
Gong, T.; Zou, B.; and Xu, Z. 2015.
\newblock Learning With $\ell_1$-Regularizer Based on Markov Resampling.
\newblock \emph{IEEE Transactions on Cybernetics}, 46(5): 1189--1201.

\bibitem[{Huber(1992)}]{huber1992robust}
Huber, P.~J. 1992.
\newblock Robust estimation of a location parameter.
\newblock In \emph{Breakthroughs in statistics}, 492--518. Springer.

\bibitem[{Khardani and Yao(2017)}]{khardani2017non}
Khardani, S.; and Yao, A.~F. 2017.
\newblock Non linear parametric mode regression.
\newblock \emph{Communications in Statistics-Theory and Methods}, 46(6):
  3006--3024.

\bibitem[{Lee(1989)}]{lee1989mode}
Lee, M.-j. 1989.
\newblock Mode regression.
\newblock \emph{Journal of Econometrics}, 42(3): 337--349.

\bibitem[{Lv, Zhu, and Yu(2014)}]{lv2014robust}
Lv, Z.; Zhu, H.; and Yu, K. 2014.
\newblock Robust variable selection for nonlinear models with diverging number
  of parameters.
\newblock \emph{Statistics \& Probability Letters}, 91: 90--97.

\bibitem[{McCracken and Ng(2016)}]{mccracken2016fred}
McCracken, M.~W.; and Ng, S. 2016.
\newblock FRED-MD: A monthly database for macroeconomic research.
\newblock \emph{Journal of Business \& Economic Statistics}, 34(4): 574--589.

\bibitem[{Meinshausen and Ridgeway(2006)}]{meinshausen2006quantile}
Meinshausen, N.; and Ridgeway, G. 2006.
\newblock Quantile regression forests.
\newblock \emph{Journal of Machine Learning Research}, 7(6).

\bibitem[{Meyn and Tweedie(2012)}]{meyn2012markov}
Meyn, S.~P.; and Tweedie, R.~L. 2012.
\newblock \emph{Markov chains and stochastic stability}.
\newblock Springer Science \& Business Media.

\bibitem[{Nikolova and Ng(2005)}]{nikolova2005analysis}
Nikolova, M.; and Ng, M.~K. 2005.
\newblock Analysis of half-quadratic minimization methods for signal and image
  recovery.
\newblock \emph{SIAM Journal on Scientific computing}, 27(3): 937--966.

\bibitem[{Paulin(2015)}]{paulin2015concentration}
Paulin, D. 2015.
\newblock Concentration inequalities for Markov chains by Marton couplings and
  spectral methods.
\newblock \emph{Electronic Journal of Probability}, 20(79): 1--32.

\bibitem[{Rudolf(2011)}]{rudolf2011explicit}
Rudolf, D. 2011.
\newblock Explicit error bounds for Markov chain Monte Carlo.
\newblock \emph{arXiv preprint arXiv:1108.3201}.

\bibitem[{Ryali et~al.(2012)Ryali, Chen, Supekar, and
  Menon}]{ryali2012estimation}
Ryali, S.; Chen, T.; Supekar, K.; and Menon, V. 2012.
\newblock Estimation of functional connectivity in fMRI data using stability
  selection-based sparse partial correlation with elastic net penalty.
\newblock \emph{NeuroImage}, 59(4): 3852--3861.

\bibitem[{Shi et~al.(2019)Shi, Huang, Feng, and Suykens}]{shi2019sparse}
Shi, L.; Huang, X.; Feng, Y.; and Suykens, J. 2019.
\newblock Sparse Kernel Regression with Coefficient-based lq-Regularization.
\newblock \emph{Journal of Machine Learning Research}, 20.

\bibitem[{Smith(2012)}]{smith2012future}
Smith, S.~M. 2012.
\newblock The future of FMRI connectivity.
\newblock \emph{Neuroimage}, 62(2): 1257--1266.

\bibitem[{Steinwart and Christmann(2008)}]{steinwart2008support}
Steinwart, I.; and Christmann, A. 2008.
\newblock \emph{Support vector machines}.
\newblock Springer Science \& Business Media.

\bibitem[{Tibshirani(1996)}]{tibshirani1996regression}
Tibshirani, R. 1996.
\newblock Regression shrinkage and selection via the lasso.
\newblock \emph{Journal of the Royal Statistical Society: Series B
  (Methodological)}, 58(1): 267--288.

\bibitem[{Wang et~al.(2017)Wang, Chen, Cai, Shen, and
  Huang}]{wang2017regularized}
Wang, X.; Chen, H.; Cai, W.; Shen, D.; and Huang, H. 2017.
\newblock Regularized modal regression with applications in cognitive
  impairment prediction.
\newblock \emph{Advances in neural information processing systems}, 30:
  1448--1458.

\bibitem[{Wu, Ying, and Zhou(2007)}]{wu2007multi}
Wu, Q.; Ying, Y.; and Zhou, D.-X. 2007.
\newblock Multi-kernel regularized classifiers.
\newblock \emph{J.Complexity}, 23(1): 108--134.

\bibitem[{Yao and Li(2014)}]{yao2014new}
Yao, W.; and Li, L. 2014.
\newblock A new regression model: modal linear regression.
\newblock \emph{Scandinavian Journal of Statistics}, 41(3): 656--671.

\bibitem[{Yao and Xiang(2016)}]{yao2016nonparametric}
Yao, W.; and Xiang, S. 2016.
\newblock Nonparametric and varying coefficient modal regression.
\newblock \emph{arXiv preprint arXiv:1602.06609}.

\bibitem[{Yu and Aristodemou(2012)}]{yu2012bayesian}
Yu, K.; and Aristodemou, K. 2012.
\newblock Bayesian mode regression.
\newblock \emph{arXiv preprint arXiv:1208.0579}.

\bibitem[{Yu, Lu, and Stander(2003)}]{yu2003quantile}
Yu, K.; Lu, Z.; and Stander, J. 2003.
\newblock Quantile regression: applications and current research areas.
\newblock \emph{Journal of the Royal Statistical Society: Series D (The
  Statistician)}, 52(3): 331--350.

\bibitem[{Zhou(2002)}]{zhou2002cover}
Zhou, D. 2002.
\newblock The covering number in learning theory.
\newblock \emph{J. Complexity}, 18: 739--767.

\bibitem[{Zhou(2003)}]{zhou2003capacity}
Zhou, D. 2003.
\newblock Capacity of reproducing kernel space in learning theory.
\newblock \emph{IEEE Trans. Inf. Theory.}, 49(7): 1743--1752.

\bibitem[{Zhou, Huang et~al.(2016)}]{zhou2016nonparametric}
Zhou, H.; Huang, X.; et~al. 2016.
\newblock Nonparametric modal regression in the presence of measurement error.
\newblock \emph{Electronic Journal of Statistics}, 10(2): 3579--3620.

\end{thebibliography}

\end{document}